\documentclass[10pt,twocolumn,letterpaper]{article}

\usepackage[pagenumbers]{cvpr} %

\usepackage{graphicx}
\usepackage{float}
\usepackage{multirow}
\usepackage{makecell}
\usepackage{pifont}
\usepackage{listings}
\usepackage{enumitem}
\usepackage{color, colortbl}
\usepackage{wrapfig}
\usepackage{algorithm}
\usepackage{algorithmic}

\newcommand{\myparagraph}[1]{\noindent{\bf #1}}

\newcolumntype{x}[1]{>{\centering\arraybackslash}p{#1pt}}
\newcolumntype{y}[1]{>{\raggedright\arraybackslash}p{#1pt}}
\newcolumntype{z}[1]{>{\raggedleft\arraybackslash}p{#1pt}}

\definecolor{mgreen}{RGB}{1,150,74}
\newcommand\up[1]{\textcolor{cvprblue}{$^{\uparrow{#1}}$}}
\newcommand\down[1]{\textcolor{red}{$^{\downarrow{#1}}$}}

\newcommand{\algname}{VisionPAD\xspace}

\definecolor{mblue}{HTML}{6895D2}
\definecolor{morange}{HTML}{E48F45}
\definecolor{mpurple}{HTML}{836096}
\definecolor{mpink}{HTML}{CD6688}
\definecolor{lightgray}{gray}{.92}

\newcommand\blfootnote[1]{%
\begingroup
\renewcommand\thefootnote{}\footnote{#1}%
\addtocounter{footnote}{-1}%
\endgroup
}

\usepackage{amssymb}  
\usepackage{bbding} 
\usepackage{pifont} 
\definecolor{cvprblue}{rgb}{0.21,0.49,0.74}
\usepackage[pagebackref,breaklinks,colorlinks,citecolor=cvprblue]{hyperref}

\definecolor{scolor}{RGB}{111,168,220}
\definecolor{hcolor}{RGB}{111,176,81}
\definecolor{ocolor}{RGB}{224,103,102}
\definecolor{wcolor}{RGB}{246,178,107}

\def\paperID{4410} %
\def\confName{CVPR}
\def\confYear{2025}

\title{\centering \includegraphics[scale=0.12, bb=-100 -100 50 0]{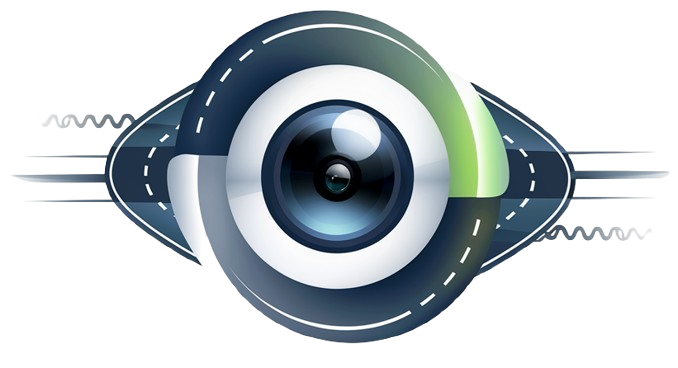}\textcolor[HTML]{5B9BD4}{Vision}\textcolor[HTML]{5B9BD4}{PAD}: A Vision-Centric Pre-training Paradigm for Autonomous Driving}

\author{
{Haiming Zhang}$^{1,2\dagger}$,
{Wending Zhou}$^{1,2\dagger}$,
{Yiyao Zhu}$^{3\dagger}$,
{Xu Yan}$^{4\textsuperscript{\ding{41}}}$\\
{Jiantao Gao}$^{4}$,
{Dongfeng Bai}$^{4}$,
{Yingjie Cai}$^{4}$,
{Bingbing Liu}$^{4}$,
{Shuguang Cui}$^{2,1}$,
{Zhen Li}$^{2,1\textsuperscript{\ding{41}}}$\\
$^1$FNii, Shenzhen~~
$^2$SSE, CUHK-Shenzhen\\
$^3$HKUST~~
$^4$Huawei Noah’s Ark Lab\\
}

\begin{document}
\maketitle

\begin{abstract}
This paper introduces VisionPAD, a novel self-supervised pre-training paradigm designed for vision-centric algorithms in autonomous driving. In contrast to previous approaches that employ neural rendering with explicit depth supervision, VisionPAD utilizes more efficient 3D Gaussian Splatting to reconstruct multi-view representations using only images as supervision.
Specifically, we introduce a self-supervised method for voxel velocity estimation. By warping voxels to adjacent frames and supervising the rendered outputs, the model effectively learns motion cues in the sequential data.
Furthermore, we adopt a multi-frame photometric consistency approach to enhance geometric perception. It projects adjacent frames to the current frame based on rendered depths and relative poses, boosting the 3D geometric representation through pure image supervision. 
Extensive experiments on autonomous driving datasets demonstrate that VisionPAD significantly improves performance in 3D object detection, occupancy prediction and map segmentation, surpassing state-of-the-art pre-training strategies by a considerable margin.
\blfootnote{$^\dagger$Work done during an internship at Huawei Noah’s Ark Lab.}
\blfootnote{$\textsuperscript{\ding{41}}$Corresponding authors.}
\end{abstract}

\section{Introduction}
\label{sec:intro}
Recent advancements in vision-centric autonomous driving, leveraging multi-view images as input~\cite{li2022bevformer,li2023bevdepth,tong2023scene,tang2024sparseocc,zhang2023radocc}, have been observed in the community due to benefits in cost efficiency, scalability, and the extensive semantic richness offered by visual inputs.
Current methodologies exhibit exceptional capability in deriving Bird's-Eye-View (BEV)~\cite{li2022bevformer,li2023bevdepth,huang2021bevdet} and occupancy features~\cite{tong2023scene,tang2024sparseocc}, achieving notable performance across various downstream applications.
Nevertheless, these models largely depend on precise 3D annotations, which poses a significant bottleneck due to the challenges involved in collecting annotations, including occupancy~\cite{wang2023openoccupancy}, 3D bounding boxes~\cite{caesar2020nuscenes}, and \etc. 

\begin{figure}[t]
    \centering
    \includegraphics[width=0.95\linewidth]{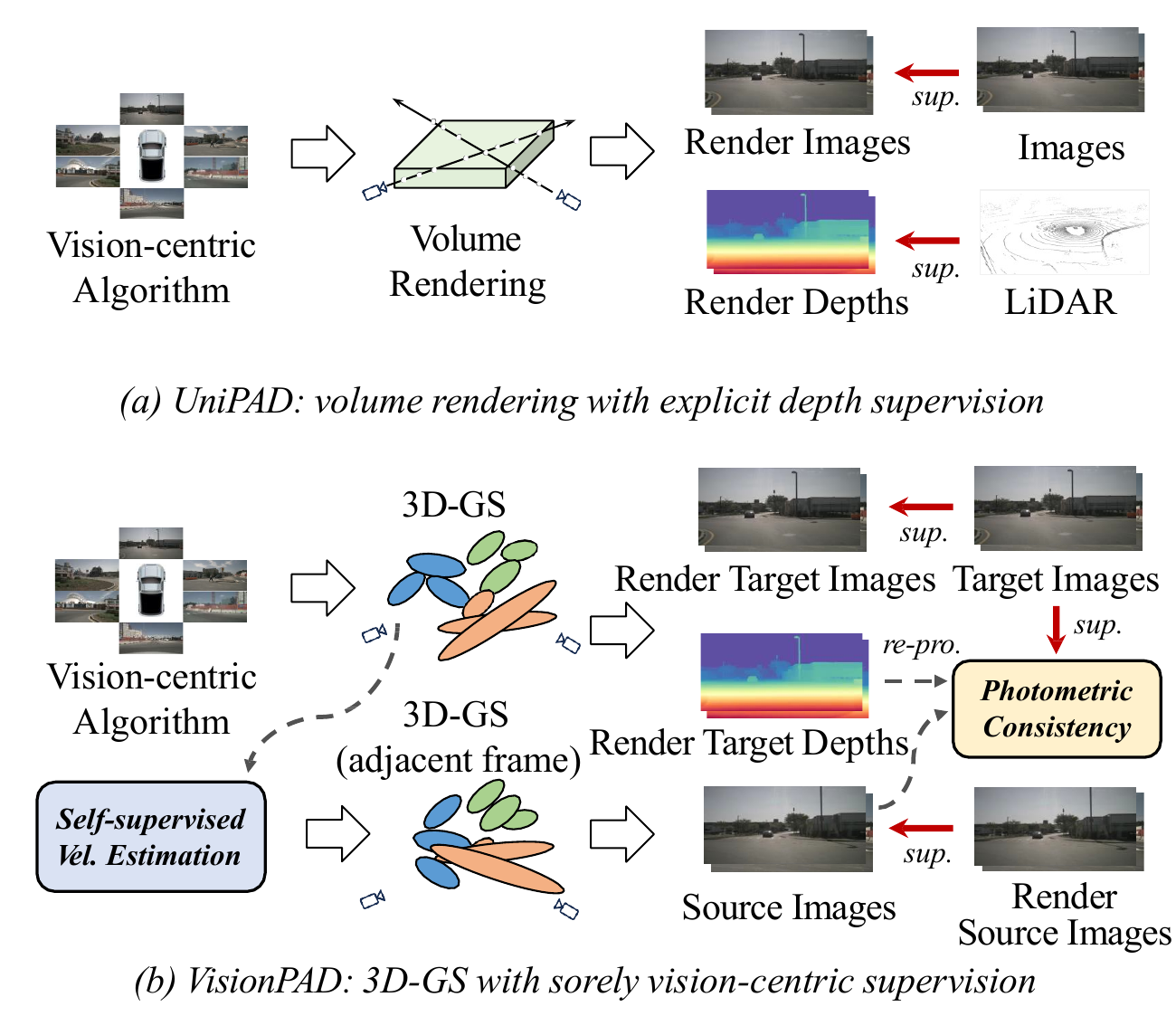}
    \caption{\textbf{Comparison with existing methods.} (a) UniPAD employs volume rendering to reconstruct the multi-view depth maps and images of the current frame, using explicit depth maps for supervision.
    (b)  In contrast, our proposed VisionPAD leverages only multi-frame, multi-view images for supervision, effectively learning motion and geometric representations through voxel velocity estimation and photometric consistency loss. re-pro. and sup. denote the re-projection and supervision, respectively.
    }

    \label{fig:teaser}
    \vspace{-0.8em}
\end{figure}

The high cost of data labeling has made pre-training a crucial strategy for scaling downstream applications in autonomous driving.
Some of the previous approaches relied on supervised pre-training for tasks like 3D object detection~\cite{park2021pseudo,wang2021fcos3d} and occupancy prediction~\cite{tong2023scene,yan2023spot}. However, these methods require large labeled datasets, which are often unavailable. 
Alternatively, other ones utilize contrastive learning~\cite{Sautier_3DV24} and Masked Autoencoders (MAE)~\cite{min2023occupancy} for self-supervised pre-training.
However, their reliance on coarse supervision makes it challenging to effectively capture semantics, 3D geometry, and temporal dynamics simultaneously~\cite{yan2024forging}.
Recently, UniPAD~\cite{yang2024unipad} introduced a pre-training paradigm by reconstructing multi-view depth maps and images from voxel features (see Fig.~\ref{fig:teaser}(a)). 
This approach uses differentiable volumetric rendering to reconstruct a complete geometric representation.
Building on this, ViDAR~\cite{yang2024vidar} incorporates next-frame prediction with a transformer and renders corresponding depth maps supervised by future LiDAR frames.
However, both techniques still rely heavily on explicit depth supervision from LiDAR data to learn 3D geometry. Relying solely on image supervision yields unsatisfactory results, limiting their applicability in camera-based autonomous driving systems.

In this paper, we propose \textbf{VisionPAD}, a self-supervised pre-training framework for vision-centric data (see Fig.~\ref{fig:teaser}(b)). Unlike prior methods that utilize volume rendering for image reconstruction, we leverage a more efficient anchor-based 3D Gaussian Splatting (3D-GS)~\cite{kerbl20233dgs}. This allows us to reconstruct higher-resolution images with the same computational budget, capturing finer-grained color details without ray-sampling~\cite{hu2023uniad}.
Furthermore, to learn motion cues solely from images, we propose a self-supervised voxel velocity estimation method. We predict per-voxel velocity using an auxiliary head and approximate voxel flow to adjacent frames using timestamps. Subsequently, we warp the current frame's voxels to adjacent frames and supervise the 3D-GS reconstruction with corresponding images.
This velocity prediction enables the model to decouple dynamic and static voxels, facilitating motion perception in downstream tasks.
Moreover, we adopt multi-frame photometric consistency loss for 3D geometric pre-training, which is a technique in self-supervised depth estimation~\cite{godard2019digging}, where the algorithm projects adjacent frames to the current frame based on rendered depths and relative poses. 

Extensive experiments on the competitive nuScenes dataset~\cite{caesar2020nuscenes}, across three downstream tasks, demonstrate the superiority of our proposed method for camera-based pre-training.
Specifically, when pre-trained solely on multi-frame image supervision, our method \textit{outperforms} state-of-the-art pre-training approaches by \textbf{+2.5 mAP} on 3D object detection, \textbf{+4.5 mIoU} on semantic occupancy prediction and \textbf{+4.1 IoU} on map segmentation.

The contributions of this paper are threefold:
\begin{itemize}
    \setlength{\itemsep}{0pt}
    \setlength{\parsep}{0pt}
    \setlength{\parskip}{0pt}
\item To the best of our knowledge, we present the first vision-centric pre-training paradigm that leverages a 3D-GS decoder for image reconstruction to improve performance on vision-centric algorithms.

\item We introduce a self-supervised voxel velocity estimation method to distinguish between static and dynamic voxel and incorporate motion information into the pre-trained model. We further adopt the photometric consistency loss to learn 3D geometric through cross-frame relative poses.

\item Our method achieves significant improvements over previous state-of-the-art pre-training approaches on three downstream tasks.
\end{itemize}

\section{Related Works}
\paragraph{Pre-training in Autonomous Driving.}
Pre-training has been extensively explored in autonomous driving to improve scalability and adaptability across diverse driving environments. Existing approaches can be broadly categorized into supervised, contrastive, masked signal modeling, and rendering-based methods.
Supervised approaches~\cite{park2021pseudo,wang2021fcos3d,tong2023scene,yan2023spot} leverage annotated driving data to learn structured representations tailored for specific tasks. 
Contrastive approaches~\cite{li2022simipu,nunes2022segcontrast,Sautier_3DV24,yuan2024ad} utilize positive and negative pairwise data to learn discriminative, view-invariant representations robust to variations within the driving scene. 
Masked signal modeling approaches~\cite{min2022voxelmae,min2023occupancy,boulch2023also,krispel2024maeli,irshad2024nerf} reconstruct occluded or masked sensory signals, fostering a holistic understanding of scene semantics.

Recent advances in rendering-based pre-training have presented novel strategies for autonomous driving perception by incorporating visual features into unified volumetric representations for rendering.
A key distinction of these rendering-based methods compared to prior work is the utilization of neural rendering~\cite{ben2020nerf}, which effectively enforces the encoding of rich geometric and appearance cues within the learned volumetric features.
UniPAD~\cite{yang2024unipad} introduces a mask generator for partial input occlusion and a modal-specific encoder to extract multimodal features in voxel space, followed by volume-based neural rendering for RGB and depth prediction. 
ViDAR~\cite{yang2024vidar} predicts future point clouds based on historical surrounding images observations.
MIM4D~\cite{mim4d} takes masked multi-frame, multi-view images as input and reconstructs the target RGB and depth using volume rendering.
BEVWorld~\cite{zhang2024bevworld} encodes image and LiDAR data into BEV tokens and utilizes NeRF rendering for reconstruction, incorporating a latent BEV sequence diffusion model to predict future BEV tokens conditioned on corresponding actions.
However, the above approaches retain a strong reliance on explicit depth supervision, inherently limiting their applicability in camera-only scenarios.

\begin{figure*}[t]
    \centering
\includegraphics[width=\linewidth]{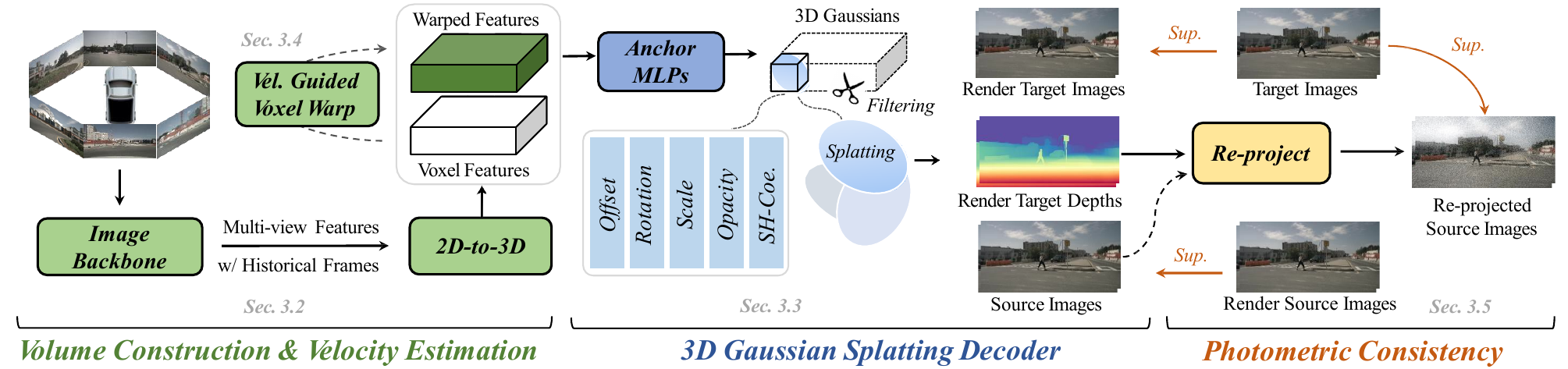}
    \caption{\textbf{Overall pipeline of VisionPAD.} 
    Taking a vision-centric perception model as the backbone, VisionPAD leverages multi-frame, multi-view images as input, generating explicit voxel representations. 
    After that, a 3D Gaussian Splatting (3DGS) Decoder reconstructs multi-view images from the voxel features. 
    After that, velocity-guided voxel warp is applied to warp current frame voxel features to adjacent frames, enabling self-supervised reconstruction via the 3D-GS Decoder using adjacent frame images as supervision. 
    Finally, a photometric consistency loss, informed by relative poses for re-projection, enforces 3D geometric constraints.
    }
    \label{fig:framework}
  \vspace{-0.8em}
\end{figure*}

\myparagraph{3D Gaussian Splatting in Autonomous Driving.}
Recently, 3D Gaussian Splatting~\cite{kerbl20233dgs} (3D-GS)-based methods have gained significant traction. 3D-GS allows for scene representation using 3D Gaussian primitives, enabling real-time rendering with minimal memory footprint via rasterization.
Several approaches have been proposed for reconstructing driving scenes using this technique. PVG~\cite{chen2023periodic} introduced Periodic Vibration Gaussians for large-scale dynamic driving scene reconstruction. DrivingGaussian~\cite{zhou2024drivinggaussian} hierarchically models complex driving scenes using sequential multi-sensor data. Furthermore, \cite{yan2024street} proposed StreetGaussians, incorporating a tracked pose optimization strategy and a 4D spherical harmonics appearance model to address the dynamics of moving vehicles. HUGS~\cite{zhou2024hugs} introduces a novel pipeline leveraging 3D Gaussian Splatting for holistic urban scene understanding. This approach entails the joint optimization of geometry, appearance, semantics, and motion using a combination of static and dynamic 3D Gaussians.
Recent works \cite{huang2024gaussianformer,gan2024gaussianocc} and \cite{chabot2024gaussianbev} utilize the 3D-GS for occucupany and BEV perception, respectively.
Nevertheless, the application of 3D-GS in pre-training for autonomous driving remains unexplored.

\section{Proposed Method}
\label{method}

This section details the proposed \textbf{VisionPAD}, outlining its core components and key innovations. 

\subsection{Overview}

The overall framework of VisionPAD is illustrated in Fig.~\ref{fig:framework}, encompassing four key modules. 
First, VisionPAD takes historical multi-frame, multi-view images and leverages a vision-centric perception network backbone with explicit representations, \ie, occupancy (Sec.~\ref{sec:voxel_rep}). 
Second, a novel 3D Gaussian Splatting Decoder reconstructs multi-view images in the current frame from voxel representations (Sec.~\ref{sec:3D-GS}). 
Third, a voxel velocity estimation strategy is proposed to predict voxel velocities, enabling the warping of current-frame voxel features to target frames. This facilitates the reconstruction of multi-view adjacent frame images and depth maps through the 3D-GS decoder. (Sec.~\ref{sec:vel_ssl}). 
Finally, leveraging the target depth maps in the current, VisionPAD incorporates 3D geometric constraints through photometric consistency loss (Sec.~\ref{sec:mpc}).
The final pre-training losses are illustrated in Sec.~\ref{sec:loss}.

\subsection{Volume Construction}
\label{sec:voxel_rep}
Given $M$ historical multi-view images $\mathbf{I}=\{I_i\}^M_{i=1}$ as input to a representative vision-centric framework, VisionPAD employs a shared image backbone to extract 2D image features, resulting in a 2D feature representation $\mathbf{F}_I\in \mathbb{R}^{N\times H\times W\times C}$, where $N$, $H$ and $W$ denote the number of views, height and width of the images, respectively.
Then, a view transformation~\cite{huang2021bevdet,li2022bevformer} lifts these 2D features into the 3D ego-centric coordinate system, generating volumetric features. 
Finally, a projection layer comprising several convolutional layers further refines the volumetric representation, which is denoted as $\mathbf{V}\in \mathbb{R}^{X \times Y \times Z \times C}$.

\subsection{3D Gaussian Splatting Decoder}
\label{sec:3D-GS}
\myparagraph{Preliminaries.}
3D Gaussian Splatting (3D-GS)~\cite{kerbl20233dgs} represents a 3D scene as a set of Gaussian primitives $\{g_k = (\mu_k, \Sigma_k, \alpha_k, c_k)\}_{k=1}^K$,  where each Gaussian $g_k$ is defined by its 3D position $\mu_k \in \mathbb{R}^3$, a covariance $\Sigma_k$, an opacity $\alpha_k\in [0, 1]$, and spherical harmonics  (SH) coefficients $c_k\in \mathbb{R}^k$.
To facilitate efficient optimization via gradient descent, the covariance matrix $\Sigma$ is generally parameterized as a product of a scaling matrix $\mathbf{S}\in \mathbb{R}^3_+$ and a rotation matrix $\mathbf{R}\in \mathbb{R}^4$:
\begin{equation}
    \Sigma = \mathbf{R S S}^T \mathbf{R}^T.
\end{equation}
Projecting 3D Gaussians onto the 2D image plane involves a view transformation $\mathbf{W}$ and the Jacobian $\mathbf{J}$ of the affine approximation of the projective transformation. The resulting 2D covariance matrix $\Sigma'$ is then given by:
\begin{equation}
    \Sigma' = \mathbf{J W \Sigma W}^T \mathbf{J}^T.
\end{equation}
To render an image from a given viewpoint, the feature of
each pixel $p$ is calculated by blending $K$ ordered Gaussians, where an alpha-blend rendering procedure~\cite{ben2020nerf} is applied:
\begin{equation}
\label{eq:alpha-blend}
    \mathbf{C}(p) = \sum_{i \in K} c_i \alpha_i \prod_{j=1}^{i-1} (1 - \alpha_i),
\end{equation}
where $K$ represents the set of Gaussians intersected by the ray corresponding to pixel $p$, and the density $\alpha_i$ is computed as the product of a 2D Gaussian with covariance $\Sigma'$ and a per-point opacity $\alpha_k$. 

To further enhance geometric representation, we incorporate multi-view depth map reconstruction, following the methodology presented in~\cite{cheng2024gaussianpro}: 
\begin{equation}
\label{eq:depth_splat}
\mathbf{D}(p) = \sum_{i \in K} d_i \alpha_i \prod_{j=1}^{i-1} (1 - \alpha_j),
\end{equation}
where $d_i$ is the distance from the $i$-th Gaussian to the camera.    
Unlike volume rendering~\cite{ben2020nerf}, 3D-GS enables efficient rendering via splat-based rasterization, projecting 3D Gaussians onto the target 2D view and rendering image patches using local 2D Gaussians.

We observe that when pre-trained solely with RGB supervision, NeRF can only sample a limited number of rays per iteration, hindering its ability to learn detailed scene geometry. 
In contrast, 3D-GS, with its efficient splat-based rasterization, exhibits a computational cost that is less sensitive to image resolution, enabling the rendering of higher-resolution images, and facilitating representation learning.

\begin{figure}[t]
    \centering
\includegraphics[width=\linewidth]{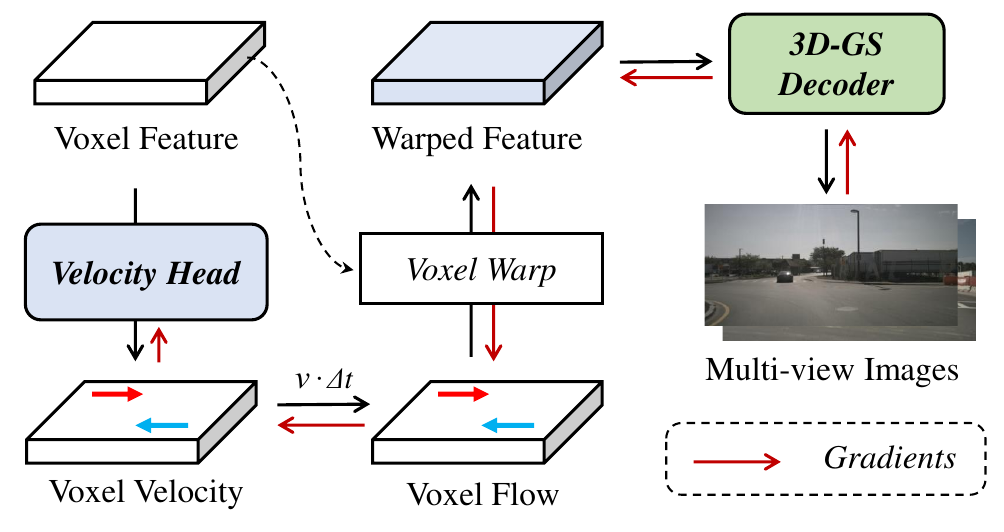}
    \caption{\textbf{Self-supervise velocity estimation.}
    Current voxel features are warped to the adjacent frame. Subsequently, multi-view images are rendered using the 3DGS Decoder and supervised by images captured in that frame.
    }
    \vspace{-1em}
    \label{fig:ssl_vel}
\end{figure}

\myparagraph{Primitives Prediction.}
Leveraging the anchor-based 3D Gaussian Splatting introduced as in~\cite{charatan2024pixelsplat}, we transform voxel features into a set of 3D Gaussians, as shown in Fig.~\ref{fig:framework}.
Each voxel center serves as an anchor point from which the attributes of multiple Gaussian primitives are predicted. 
Specifically, we employ MLPs to regress the parameters of each Gaussian on the anchor (\ie, anchor MLPs), including its offset from the voxel center, spherical harmonic coefficients, opacity, scale, and rotation.
This representation facilitates efficient multi-view image synthesis via differentiable rendering, utilizing known camera intrinsics and extrinsic through Eqn.~\ref{eq:alpha-blend}. 
Finally, the 3D-GS decoder generates multi-view images  $\mathbf{C}\in \mathbb{R}^{N\times H\times W\times 3}$ which are supervised by multi-view images in the current frame, where $N$ is the number of views, and $H$ and $W$ are the height and width of the images, respectively.

\myparagraph{Gaussian Filtering.}
To reduce computational overhead during pre-training, we filter low-confidence Gaussians based on their predicted opacity. Specifically, the opacity is predicted using a tanh activation function, and Gaussians with predicted opacities less than 0 are discarded.

\subsection{Self-supervised Voxel Velocity Estimation}
\label{sec:vel_ssl}

In this paper, we introduce a self-supervised approach for estimating per-voxel velocities, leveraging the inherent temporal consistency of objects within a scene. 
This motion information enriches the representation and facilitates understanding of the dynamic scene. 

\myparagraph{Velocity Guided Voxel Warping.}
As depicted in Fig.~\ref{fig:ssl_vel}, an auxiliary velocity head, appended to the voxel features, regresses per-voxel velocity vectors in the world coordinates. 
After that, we approximate the per voxel flow by scaling the predicted velocities by the inter-frame time interval, enabling the transformation of voxel features from the current frame to an adjacent frame. 
Subsequently, we warp the voxel features to their estimated positions in the adjacent frame.
The above process can be easily implemented through \textit{GridSample} operator.

\myparagraph{Adjacent Frame Rendering.}
After obtaining the warped voxel features for the adjacent frame, we employ the 3D-GS decoder (Sec.~\ref{sec:3D-GS}) to render multi-view images. These rendered images are then compared against the corresponding ground truth images of the adjacent frame using a supervised loss. 
It should be noted that, during backpropagation, only the parameters of the velocity head are updated. This targeted optimization strategy encourages the network to prioritize learning discriminative motion features, leading to improved performance.

\begin{table*}[t]
	\centering
	
        \begin{tabular}{l|c|c|c|c|cc}
		\toprule
		Methods & Venue & Pre-train Modal & CS & CBGS & NDS (\%) $\uparrow$ & mAP (\%) $\uparrow$ \\
		\midrule
		BEVFormer-S~\cite{li2022bevformer} & ECCV'22 & - & & \checkmark & 44.8 & 37.5 \\
		SpatialDETR~\cite{doll2022spatialdetr} & ECCV'22 & - & & & 42.5 & 35.1 \\
		PETR~\cite{liu2022petr} & ECCV'22 & - & & \checkmark & 44.2 & 37.0 \\
		Ego3RT~\cite{lu2022ego3rt} & ECCV'22 & - & & & 45.0 & 37.5 \\
		3DPPE~\cite{shu20233dppe} & ICCV'23 & - & & \checkmark & 45.8 & 39.1 \\
            BEVFormerV2~\cite{yang2023bevformerv2} & CVPR'23 & - & & & 46.7 & 39.6 \\
		CMT-C~\cite{yan2023cmt} & ICCV'23 & -  & & \checkmark & 46.0 & 40.6 \\
		FCOS3D & ICCVW'21 & - & & & 38.4 & 31.1 \\
		\hline
		UVTR~\cite{li2022uvtr} & NeurIPS'22  & - & & & 45.0 & 37.2 \\
        
		{UVTR+UniPAD}$^\dagger$~\cite{yang2024unipad} & CVPR'24 & C & & & ~~~~~~~44.8~\down{0.2} & ~~~~~~~38.5~\up{1.3} \\
        \rowcolor[gray]{.92}
        \textbf{UVTR+VisionPAD (Ours)} & - & C & & & ~~~~~~~\textbf{46.7}~\up{1.7} & ~~~~~~~\textbf{41.0}~\up{3.8} \\\hline

		UVTR~\cite{li2022uvtr} & NeurIPS'22  & - &\checkmark & & 48.8 & 39.2 \\
        {UVTR+UniPAD}$^\dagger$~\cite{yang2024unipad} & CVPR'24 & C &\checkmark & & ~~~~~~~48.6~\down{0.2} & ~~~~~~~40.5~\up{0.7} \\

        {UVTR+UniPAD} & CVPR'24 & C+L &\checkmark & & ~~~~~~~50.2~\up{1.4} & ~~~~~~~42.8~\up{3.6} \\

        \rowcolor[gray]{.92}
        \textbf{UVTR+VisionPAD (Ours)} & - & C &\checkmark & & ~~~~~~~{49.7}~\up{0.9} & ~~~~~~~{41.2}~\up{2.0} \\
        \rowcolor[gray]{.92}
        \textbf{UVTR+VisionPAD (Ours)} & - & C+L & \checkmark& & ~~~~~~~\textbf{50.4}~\up{1.6} & ~~~~~~~\textbf{43.1}~\up{3.9} \\
		\bottomrule
	\end{tabular}%
    \caption{
		\textbf{3D object detection performance on the nuScenes {\em val} set.}
        Our approach achieves superior performance among existing vision-centric methods. ``CS" denotes camera sweeps (two historical frames) input. ``CBGS" refers to the class-balanced grouping and sampling~\cite{zhu2019class}. 
        $\dagger$ indicates our implementation. Pre-train Modal denotes the supervision modalities used during the pre-training stage.
	}
	\label{tab:object_det}
  \vspace{-1.0em}
\end{table*}

\subsection{Photometric Consistency}
\label{sec:mpc}
Photometric consistency is proposed for self-supervised depth estimation~\cite{godard2019digging}. It leverages the predicted depth maps of a target frame $I_t$ to re-project source frames $I_{t'}$ into the source view:
\begin{equation}
\mathbf{I}_{t^{\prime}\rightarrow t} = \mathbf{I}_{t^{\prime}} \left< \mathrm{proj}(\mathbf{D}_t, \mathbf{T}_{t\rightarrow t^{\prime}}, \mathbf{K})\right>,
\label{eq:depth_ct}
\end{equation}
where $\left< \cdot \right>$ denotes the differentiable grid sampling operator, $\mathbf{D}_t$ represents the predicted depth map for the target frame, $\mathbf{T}_{t\rightarrow t^{\prime}}$ is the relative pose transforming points from frame $t$ to $t'$, and $\mathbf{K}$ represents the camera intrinsics.
$\mathrm{proj}(\cdot)$ computes the 2D pixel coordinates in the source frame $\mathbf{I}_{t^{\prime}}$ corresponding to the projected depth values $\mathbf{D}_t$. The photometric consistency loss is then computed as:
\begin{equation}
  \mathcal{L}_{pc} = \alpha(1-\mathrm{SSIM}(\mathbf{I}_t,\mathbf{I}_{t^{\prime}\rightarrow t})) + (1-\alpha)\Vert \mathbf{I}_t - \mathbf{I}_{t^{\prime}\rightarrow t} \Vert_1,
  \label{eq:pc}
\end{equation}
where $\mathbf{I}_t$ and $\mathbf{I}_{t^{\prime}\rightarrow t}$ are the target frame and re-projected images, respectively. $\alpha$ is a weighting hyperparameter balancing the SSIM and $L_1$ terms.

To enhance the geometric representation, we incorporate photometric consistency during pre-training by utilizing the rendered depth map from the current frame as $D_t$ in Eqn.~\ref{eq:depth_ct}. This depth map, generated via the 3D Gaussian Splatting decoder (Sec.~\ref{sec:3D-GS}), can be expressed as:
\begin{equation}
  \mathbf{D}_t = \mathrm{3DGS}(\mathbf{V}_t, \mathbf{K}_t, \mathbf{T}_t),
  \label{eq:depth_pred}
\end{equation}
where $\mathbf{V}_t$, $\mathbf{K}_t$ and $\mathbf{T}_t$ are volume features, camera intrinsics and extrinsics, respectively.

\subsection{Pre-training Loss}
\label{sec:loss}
Our pre-training strategy employs three loss terms. First, we apply an L1 reconstruction loss $\mathcal{L}_{img}$ to the multi-view images rendered from the current frame's voxel features via the 3D-GS decoder, comparing them against the corresponding ground truth images.
Second, the self-supervised velocity estimation is supervised by an L1 loss $\mathcal{L}_{vel}$ applied to multi-view images rendered from the warped voxel features. This loss encourages accurate velocity prediction by ensuring consistent reconstructions after motion compensation. 
Finally, we incorporate photometric consistency loss $\mathcal{L}_{pc}$, further refining the model's understanding of scene geometry. 
This combination of losses can be formulated as:
\begin{equation}
  \mathcal{L} = \omega_1 \mathcal{L}_{img} + \omega_2 \mathcal{L}_{vel} + \omega_3 \mathcal{L}_{pc},
  \label{eq:loss}
\end{equation}
where $\omega_1$, $\omega_2$ and $\omega_3$ are 0.5, 1, 1, respectively.

\begin{table*}[t]
  \centering
  \scalebox{1}{
  \begin{tabular}{l|c|ccc|c}
    \toprule[1.5pt]
    Methods & Venue & Backbone & Image Size & Pre-train by Det. & mIoU (\%) $\uparrow$ \\
    \midrule
    BEVFormer \cite{li2022bevformer} & ECCV'22 & R101 & $900\times 1600$ & \checkmark & 39.3 \\
    TPVFormer \cite{huang2023tri} & CVPR'23& R50 & $900\times 1600$ & \checkmark& 34.2 \\
    
    FB-Occ (16f)~\cite{li2023fb} & CVPRW'23 &  R50 & $384\times 704$ & \checkmark & 39.1 \\
    RenderOcc~\cite{pan2024renderocc} & ICRA'24 & Swin-B & $512\times 1408$ & \checkmark & 24.5 \\
    SparseOcc (16f)~\cite{liu2024fully} & ECCV'24 & R50 & $256\times 704$ & - & 30.6 \\
    OPUS (8f)~\cite{wang2024opus} & NeurIPS'24 & R50 &  $256\times 704$ & - & 36.2 \\
    \midrule
    UVTR$^\dagger$~\cite{li2022uvtr} & NeurIPS'22 & ConvNeXt-S & $900\times 1600$ & -& 30.1\\
    
    UVTR+UniPAD$^\dagger$~\cite{yang2024unipad} & CVPR'24 & ConvNeXt-S & $900\times 1600$ & - & ~~~~~~~31.0~\up{0.9}\\

    \rowcolor[gray]{.92}
    \textbf{UVTR+VisionPAD (Ours)} & - & ConvNeXt-S & $900\times 1600$ & - & ~~~~~~~\textbf{35.4}~\up{5.4}\\\hline

    BEVDet-Occ (8f)~\cite{huang2021bevdet} & ArXiv'22 &  R50 & $384\times 704$ & \checkmark & 39.3 \\
    \rowcolor[gray]{.92}
    \textbf{BEVDet-Occ (8f)+VisionPAD (Ours)} & - &  R50 & $384\times 704$ & \checkmark & ~~~~~~~\textbf{42.0}~\up{2.7}\\
    \bottomrule[1.5pt]
  \end{tabular}
  }
  \caption{\textbf{Semantic Occupancy prediction performance on Occ3D \textit{val} set.} ``8f" and ``16f" indicate models incorporating temporal information from 8 and 16 frames, respectively. Results are largely sourced from OPUS~\cite{wang2024opus}, with those marked $^\dagger$ indicating our own implementation. ``Pre-train by Det." refers to methods initializing with weights pre-trained for 3D object detection.
  }
  \label{tab:occ}
  \vspace{-1.0em}
\end{table*}

\begin{table}[t]
  \centering
  \scalebox{1}{
  \begin{tabular}{l|c|c}
    \toprule[1.5pt]
    Methods  & Backbone  & Lanes (\%) $\uparrow$ \\ %
    \midrule
    UVTR~\cite{li2022uvtr} & ConvNeXt-S  & 15.0 \\

    UVTR+UniPAD~\cite{yang2024unipad} & ConvNeXt-S  & ~~~~~~~16.3~\up{1.3} \\

    \rowcolor[gray]{.92}
    \textbf{UVTR+Ours} & ConvNeXt-S & ~~~~~~~\textbf{20.4}~\up{5.4} \\
    
    \bottomrule[1.5pt]
  \end{tabular}
  }
  \caption{\textbf{Map segmentation performance.} All reported results utilize the map decoder from UniAD~\cite{hu2023uniad}, with BEV segmentation lane Intersection over Union (IoU) as the evaluation metric.
  }
  \vspace{-1em}
  \label{tab:map_seg}
\end{table}

\section{Experiments}

\subsection{Experimental Settings}
\myparagraph{Dataset and Metrics.} 
We evaluate our approach on the nuScenes dataset~\cite{caesar2020nuscenes}, a large-scale autonomous driving benchmark comprising 700 training, 150 validation, and 150 testing scenes.
Each scene provides synchronized six-camera surround-view images and LiDAR point clouds, capturing rich 3D information in diverse urban environments. 
The dataset features comprehensive annotations for tasks such as 3D object detection and 3D map segmentation. 
The annotations of semantic occupancy prediction are provided by \cite{tian2024occ3d}.

We employ the standard nuScenes Detection Score (NDS) and mean Average Precision (mAP) for evaluating the 3D object detection task. 
Semantic occupancy prediction and map segmentation performance are assessed using mean Intersection-over-Union (mIoU) and Intersection-over-Union (IoU), respectively.

\myparagraph{Implementation Details.}
Our implementation is based on MMDetection3D~\cite{mmdet3d2020}. 
Following the same configuration as UniPAD~\cite{yang2024unipad}, we use a ConvNeXt-small~\cite{liu2022convnext} as the default image encoder with an input resolution of $1600 \times 900$ pixels.
The constructed volume has shape $180\times 180 \times 5$ within a valid perception range of [-54m, 54m] for the $X$ and $Y$ axis, and [-5m, 3m] for the $Z$ axis. 
The channel number is set to 256 for volume features.
Data augmentation during pre-training includes random scaling and rotation, along with partial input masking (size=32, ratio=0.3 for images). 
The pre-training decoder consists of a 2-layer MLP and a 3D Convolution-BatchNorm-ReLU refine network for the velocity head, another 2-layer MLP with ReLU as an activation layer between them for predicting Gaussian primitives. 
The model is pre-trained for $12$ epochs using the AdamW optimizer, with initial learning rates set to $2 \times 10^{-4}$, total batch size is 4.
Fine-tuning follows the official downstream model configurations without modification. 

\subsection{Main Results}
\begin{figure*}[t]
    \centering
\includegraphics[width=1\linewidth]{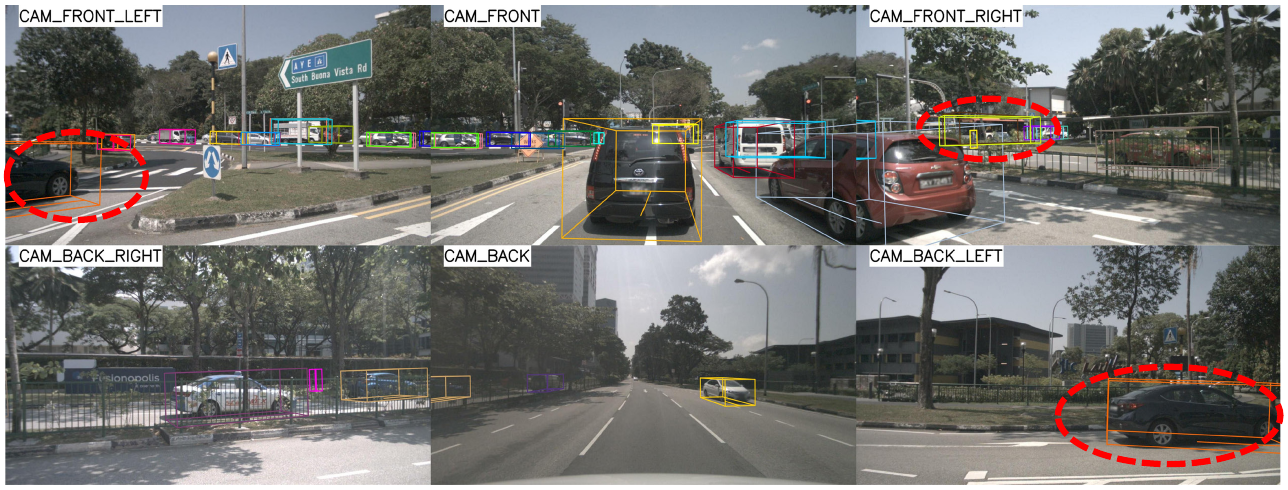} %
\includegraphics[width=1\linewidth]{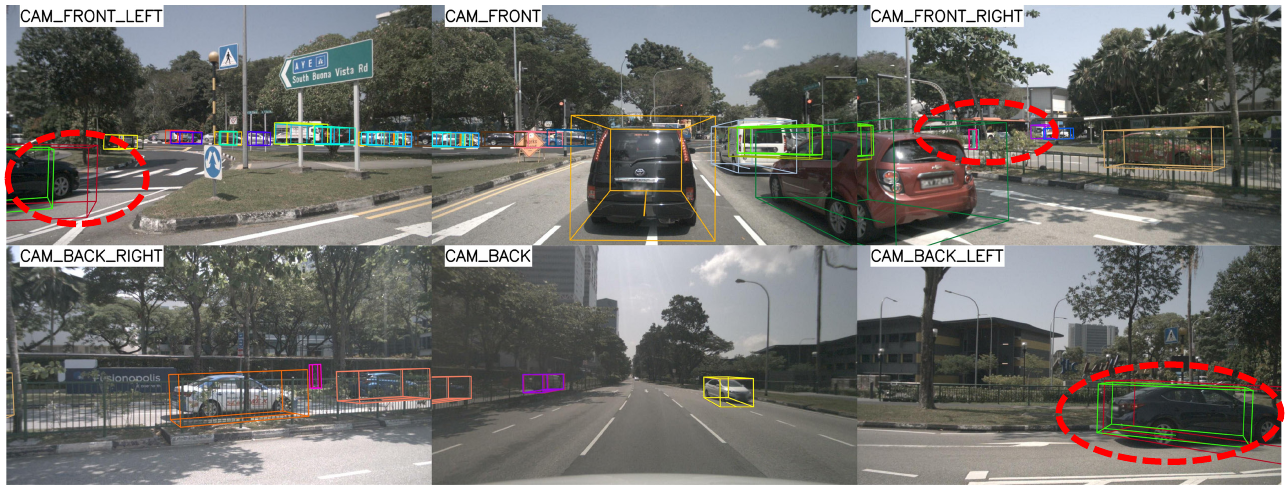} %
    \caption{\textbf{Qualitative comparison of 3D object detection between VisionPAD (top) and UniPAD (bottom) on nuScenes \textit{val} set.} Each predicted object instance is illustrated by a unique colored 3D bounding box. VisionPAD demonstrably mitigates both false positive and false negative detections (highlighted within red circles).}
    \label{fig:visualization}
    \vspace{-1.0em}
\end{figure*}

We evaluate the effectiveness of VisionPAD on three challenging downstream perception tasks: 3D object detection, semantic occupancy prediction, and map segmentation. The visualization of 3D object detection can be found in Fig.~\ref{fig:visualization}.

\myparagraph{3D Object Detection.} 
As shown in Tab.~\ref{tab:object_det}, we benchmark VisionPAD against state-of-the-art vision-centric 3D object detection methods on the nuScenes validation set. 
Using UVTR~\cite{li2022uvtr} as our baseline, we report the performance achieved with UniPAD pre-training.
Despite being trained exclusively on image data, VisionPAD demonstrates significant improvements over the baseline, achieving gains of 1.7 NDS and 3.8 mAP without multi-frame as input. 
Remarkably, without test-time augmentation, model ensembling or LiDAR supervision, VisionPAD attains \textbf{49.7} NDS and \textbf{41.2} mAP, surpassing existing state-of-the-art approaches with historical frames.

In contrast, under the same supervisory signals, UniPAD pre-training offers only marginal improvements over the baseline, even exhibiting a slight decrease in NDS from 45.0 to 44.8. 
While incorporating LiDAR-projected depth maps as additional supervision does improve the performance, this highlights the limitations of relying solely on geometric supervision derived from RGB images, which hinders generalization. 
VisionPAD, with its strong explicit depth supervision, consistently outperforms UniPAD, demonstrating the effectiveness of our proposed approach.

\begin{figure}[t]
    \centering
\includegraphics[width=1.0\linewidth]{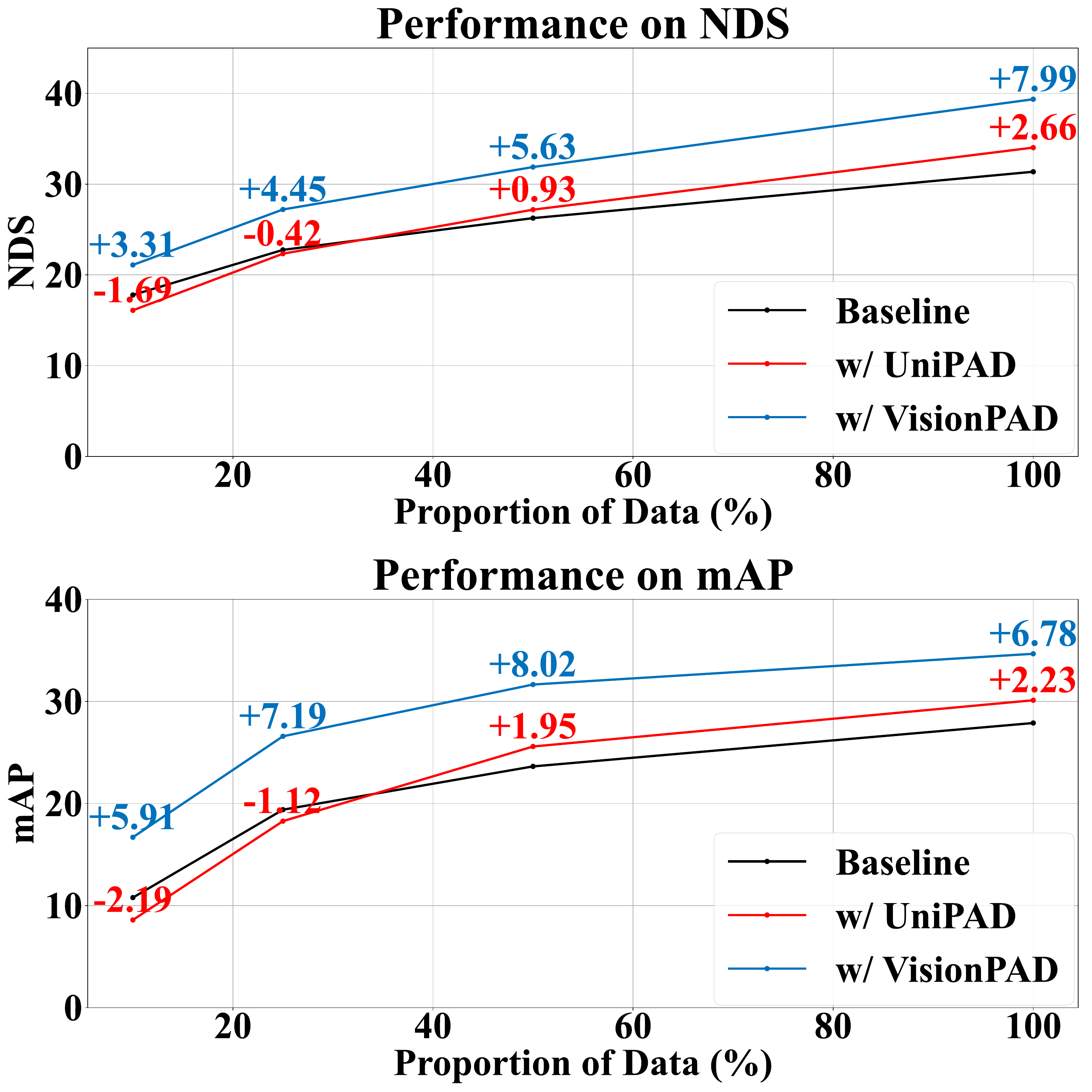}
    \vspace{-1.5em}
    \caption{\textbf{Data efficiency with limited data.} We evaluate VisionPAD's data efficiency by reducing the proportion of available annotations used during downstream fine-tuning for 3D object detection. Results highlight the effectiveness of our pre-training.
    }
    \label{fig:data_effi}
    \vspace{-1.5em}
\end{figure}

\myparagraph{Semantic Occupancy Prediction.}
As demonstrated in Tab.~\ref{tab:occ}, our image-modality-only pre-training method surpasses certain 3D detection supervised pre-training approaches, achieving 35.4\%, compared to 34.2\% for TPVFormer~\cite{huang2023tri}. 
Furthermore, its performance is comparable even to methods leveraging long temporal clues. Under the same settings, our pre-training technique significantly improves mIoU performance, from 30.1\% in UVTR to 35.4\%. Notably, without depth supervision during pre-training, UniPAD achieves only a marginal 1\% mIoU improvement. 
Even with a stronger baseline (\ie, BEVDet~\cite{huang2021bevdet}) using seven historical frames as input, pre-training with VisionPAD still yields a notable improvement in performance, from 39.3\% to \textbf{42.0\%}.
These results highlight the effectiveness of VisionPAD for dense occupancy prediction.

\begin{table*}[t]
\centering

\begin{tabular}{l|c|cc|cc|cc} 
\toprule
Methods & Vol. Rend. & 3DGS Dec. & Gaus. Filter  & V.V. Est. & P.C. &  NDS & mAP \\ 
\midrule
UVTR (baseline)~\cite{li2022uvtr} &  &  &  &   &  & 22.8 & 19.4 \\
UniPAD~\cite{yang2024unipad} & \checkmark &  &  &  &   & ~~~~~~~22.3~\down{0.5} &  ~~~~~~~18.3~\down{1.1} \\
\midrule
Model A &  & \checkmark &  &  &   & ~~~~~~~22.8~\up{0.0} & ~~~~~~~18.2~\down{1.2}  \\
Model B &  & \checkmark & \checkmark &  &   & ~~~~~~~23.4~\up{0.6} & ~~~~~~~18.9~\down{0.5} \\
Model C &  & \checkmark & \checkmark  & \checkmark &  & ~~~~~~~23.6~\up{0.8} & ~~~~~~~20.1~\up{0.7} \\
Model D &  & \checkmark & \checkmark  &  & \checkmark   & ~~~~~~~26.0~\up{3.2} & ~~~~~~~24.5~\up{5.1} \\
\rowcolor[gray]{.92}
\textbf{VisionPAD (Ours)} &  & \checkmark & \checkmark & \checkmark & \checkmark &  ~~~~~~~\textbf{27.3}~\up{4.5} & ~~~~~~~\textbf{26.5}~\up{7.1} \\
\bottomrule
\end{tabular}
\caption{\textbf{Ablation studies.} We report the NDS and mAP metrics in the nuScenes \textit{val} set for the 3D object detection task. ``Dec.'', ``V.V. Est'' and ``P.C.'' denote decoder, voxel velocity estimation and photometric consistency, respectively.}
\vspace{-1em}
\label{tab:abl}
\end{table*}

\myparagraph{Map Segmentation.}
Tab.~\ref{tab:map_seg} presents the performance gains of our method and UniPAD~\cite{yang2024unipad} on the map segmentation task. 
We utilize UVTR~\cite{li2022uvtr} as our baseline and employ the map decoder from UniAD~\cite{hu2023uniad} for map segmentation prediction. 
The results demonstrate that UniPAD offers only a limited improvement over UVTR (+1.3\%), while VisionPAD substantially boosts performance by 5.4\%.

\subsection{Comprehensive Analysis}
To reduce training time, we adopt a \textbf{lighter backbone model} following UniPAD~\cite{yang2024unipad}  for data efficiency and ablation study experiments. Specifically, we increase the voxel size of the volume feature from $0.6$ to $0.8$, while keeping the z-axis resolution unchanged, resulting in a volume feature with $128\times 128 \times 5$ shape. The channel dimension of intermediate features is also reduced from 256 to 128.
For the ablation study, we use 50\% data for pre-training and 25\% for fine-tuning.

\myparagraph{Data Efficiency.} 
A crucial benefit of pre-training is its capacity to enhance data efficiency in downstream tasks with limited labeled data. 
To further validate the benefits of our pre-training approach when pre-training data is abundant but downstream data is scarce, we conduct experiments using varying proportions of the nuScenes training set (10\%, 25\%, 50\%, and 100\%) to fine-tune a model pre-trained on the full nuScenes training dataset within the same epoches.

Fig.~\ref{fig:data_effi} illustrates the data efficiency benefits of VisionPAD. When using only image supervision for pre-training, VisionPAD, trained and fine-tuned on the complete dataset, outperforms UniPAD by +5.3 NDS and +4.5 mAP. This advantage becomes even more pronounced with decreasing amounts of fine-tuning data. 
For example, when fine-tuning on 50\% and 25\% of the data, the mAP improvement increases to approximately +6 mAP. 
These results underscore the effectiveness of VisionPAD in leveraging purely visual supervision for substantial performance gains, particularly in data-scarce regimes. 

\myparagraph{Ablation Study.}
Tab.~\ref{tab:abl} presents an ablation study to dissect the contributions of each component in our proposed model. 
The upper section establishes the performance baseline with UVTR~\cite{li2022uvtr} and the results obtained using UniPAD~\cite{yang2024unipad}. We observe that image-only pre-training with UniPAD leads to a performance decrease in both NDS and mAP during fine-tuning.
This suggests an inability to fully exploit the information contained within multi-view image supervision.

The lower section details the impact of our proposed modifications. Model A replaces UniPAD's volume rendering with our 3DGS decoder. The 3DGS decoder, leveraging full-image rendering and higher-resolution supervision, exhibits superior fine-tuning performance compared to volume rendering. 
Model B introduces filtering of low-confidence Gaussians, which effectively improves pre-training performance. Using only single-frame surround-view images for supervised pre-training, this model achieves a +0.6 NDS gain during fine-tuning. 
Model C incorporates voxel velocity estimation (V. V. Est.), resulting in a +1.2 mAP improvement in downstream fine-tuning.
The inclusion of a photometric consistency loss (P.C.) in Model D significantly boosts pre-training performance, yielding gains of +2.4 NDS and +4.4 mAP compared to Model C. 
Finally, our complete model, \textbf{VisionPAD}, incorporating the photometric consistency loss with voxel velocity estimation, achieves substantial improvements over the baseline, with gains of \textbf{+4.5 NDS} and \textbf{+7.1 mAP}. This demonstrates the effectiveness of our proposed components.

Additionally, we found that using 2/3/4 anchors per voxel does not improve performance, with NDS dropping by 0.1/0.5/0.4.

\begin{figure}[t]
    \centering
\includegraphics[width=\linewidth]{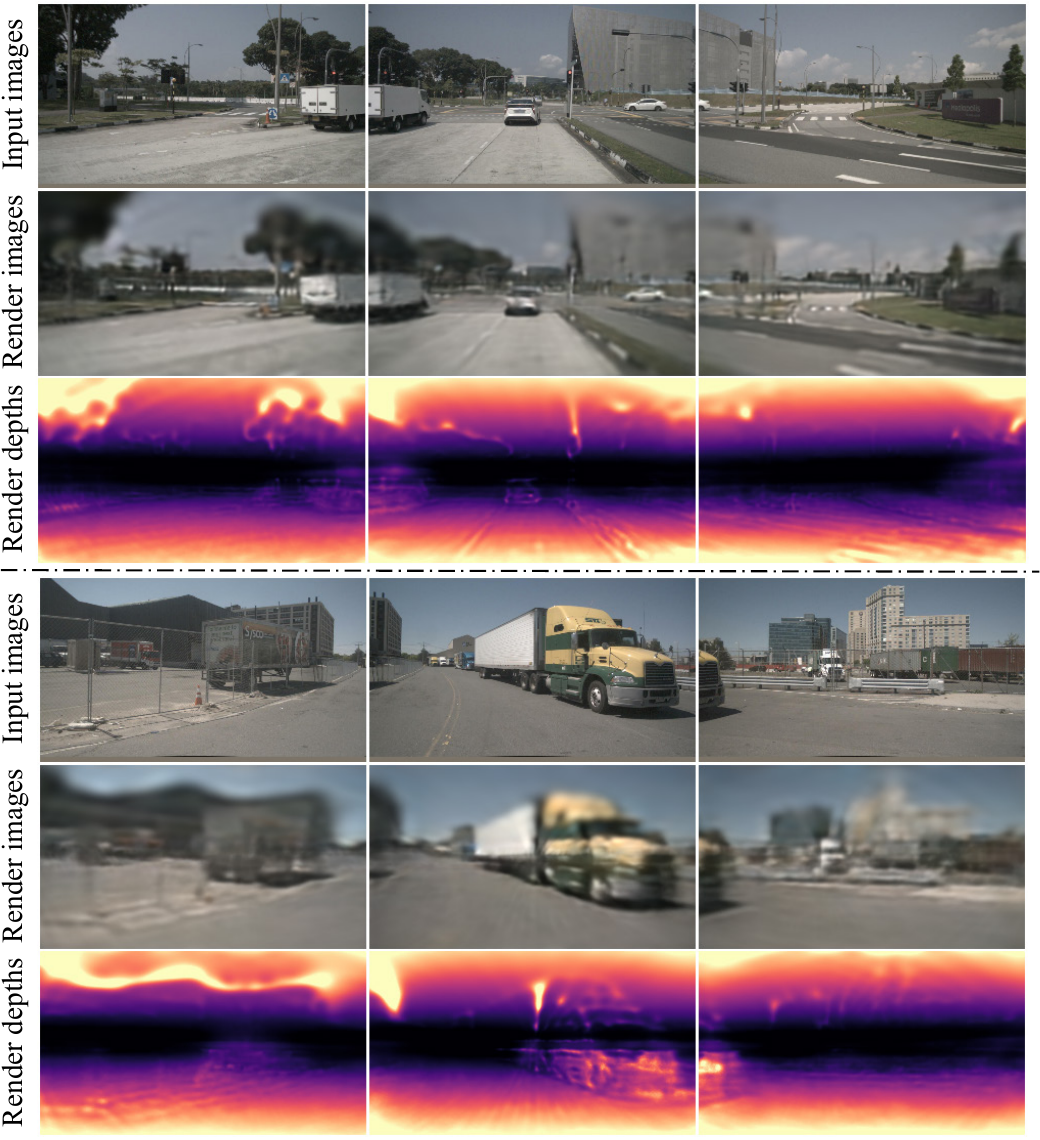}
    \caption{\textbf{Rendering results.} Leveraging multi-view images as supervision, VisionPAD demonstrates compelling depth and image reconstruction after pre-training.
    }
    \label{fig:render_vis}
    \vspace{-1em}
\end{figure}

\myparagraph{Visualization of Rendering.}
As shown in Fig.~\ref{fig:render_vis}, employing solely multi-view images for self-supervised pre-training, VisionPAD exhibits promising depth and image reconstruction capabilities.

\section{Conclusion}
This paper introduces VisionPAD, a novel self-supervised pre-training framework for vision-centric autonomous driving. Leveraging Gaussian Splatting, self-supervised voxel velocity estimation, and photometric consistency approach, VisionPAD eliminates the reliance on explicit depth supervision. Our results demonstrate improved performance and reduced computational overhead compared to existing neural rendering-based methods requiring depth supervision. This work establishes a new paradigm for efficient and scalable self-supervised vision-centric pre-training.

\section*{Acknowledgements}
This work was supported by Shenzhen General Program No. JCYJ20220530143600001, by the Basic Research Project No. HZQB-KCZYZ-2021067 of Hetao Shenzhen HK S\&T Cooperation Zone, by NSFC with Grant No. 62293482, by Shenzhen-Hong Kong Joint Funding No. SGDX20211123112401002, by the Shenzhen Outstanding Talents Training Fund 202002, by Guangdong Research Project No. 2017ZT07X152 and No. 2019CX01X104, by the Guangdong Provincial Key Laboratory of Future Networks of Intelligence (Grant No. 2022B1212010001), by the Guangdong Provincial Key Laboratory of Big Data Computing, CHUK-Shenzhen, by the NSFC 61931024\&12326610, by the Key Area R\&D Program of Guangdong Province with grant No. 2018B030338001, by the Shenzhen Key Laboratory of Big Data and Artificial Intelligence (Grant No. ZDSYS201707251409055), by Tencent \& Huawei Open Fund, and by China Association for Science and Technology Youth Care Program.

{
    \small
    \bibliographystyle{ieeenat_fullname}
    \bibliography{main}
}

\clearpage

\appendix
\maketitle
\noindent\textbf{\Large Appendix}

\setcounter{page}{1}

\renewcommand{\thesection}{\Alph{section}}

\section{Additional Implementation Details}
In this section, we elaborate on the implementation details of our proposed approach.

\subsection{Evaluation Metrics}
\myparagraph{3D Objection Detection.} \label{sec:obj_det}
In accordance with the evaluation protocols established in previous works~\cite{li2022bevformer,huang2021bevdet,yang2024unipad}, we utilize the NuScenes Detection Score (NDS) and mean Average Precision (mAP) as the primary metrics for assessing 3D object detection performance. 
The mAP is computed using the center distance on the ground plane rather than the 3D Intersection over Union (IoU), to match the predicted results and ground truth. 
Furthermore, the nuScenes dataset incorporates five types of true positive metrics (TP metrics), including ATE, ASE, AOE, AVE, and AAE for measuring translation, scale, orientation, velocity, and attribute errors, respectively. 
These TP metrics are defined for each class, and their means are computed across classes to yield mATE, mASE, mAOE, mAVE, and mAAE.
Based on these TP errors, we define the TP score as $\mathrm{TP}_{\mathrm{score}}=\mathrm{max}(1-\mathrm{TP}_{\mathrm{error}}, 0)$.
Subsequently, the nuScenes detection score (NDS) is computed as:
\begin{align}
    \mathrm{NDS} = \frac{1}{10}\left[5\mathrm{mAP}+\sum{\mathrm{TP}_{\mathrm{score}}}\right],
\end{align}
to encapsulate all aspects of the nuScenes detection comprehensively.

\myparagraph{Semantic Occupancy Prediction.} \label{sec:occ_pred}
In the context of 3D semantic occupancy prediction, we adhere to the evaluation methodology employed by existing methods~\cite{tian2024occ3d,li2023fb,huang2023tri}, utilizing mean intersection-over-union (mIoU) across all classes and intersection-over-union (IoU) as the evaluation metrics:
\begin{align}
    \mathrm{mIoU}&=\frac{1}{|C'|} \sum_{i \in C'} \frac{{TP}_i}{{TP}_i+{FP}_i+{FN}_i},
\end{align}
where $C'=17$, $c_0$, $TP$, $FP$, $FN$ denote the number of semantic classes in the Occ3D-nuScenes dataset~\cite{tian2024occ3d}, the empty class, the number of true positive, false positive and false negative predictions, respectively.

\begin{algorithm}[t]
	\caption{The pseudocode of voxel velocity estimation.}          
	\label{alg:flow}                        
	\begin{algorithmic}
		\renewcommand{\algorithmicrequire}{\textbf{Input:}}
		\renewcommand{\algorithmicensure}{\textbf{Output:}}
		\REQUIRE $\mathcal{V}_t$, $M$    
		\ENSURE$\mathcal{V}_{t+n}$
		\STATE \textit{\# predict the absolute flow for each voxel}
		\STATE  $\mathcal{F}_{t} \leftarrow \textbf{flow\_decoder}(\mathcal{V}_t)$
		\STATE \textit{\# transform the flow into absolute grid displacement to the future}
        \STATE $\hat{\mathcal{F}}_{t+n} \leftarrow \Delta{t} \cdot n \cdot \mathcal{F}_f$
		\STATE \textit{\# transform the absolute displacement to the future ego coordinate by using the current to future transformation}
		\STATE $\mathcal{\tilde{F}}_{t+n} \leftarrow M \cdot \mathcal{\hat{F}}_{t+n}$   
		\STATE \textit{\# warping the current frame volume feature to the future}
		\STATE $\mathcal{V}_{t+n} \leftarrow \textbf{grid\_sample}(\mathcal{V}_t,~\mathcal{\tilde{F}}_{t+n}$)

		\RETURN $\mathcal{V}_{t+n}$
	\end{algorithmic}
\end{algorithm}

\subsection{Details of Voxel Velocity Estimation Module}
In the main paper, we have presented the effectiveness of the proposed self-supervised voxel velocity estimation module in capturing inherent temporal consistencies. 
Here we elaborate on more details of this module.
The pseudocode for the voxel velocity estimation module is presented in Alg.~\ref{alg:flow}.
In this context, the inputs $\mathcal{V}_t$, $M$ represent the extracted volume features at the current frame and the pose transformation matrix from the current frame to a future frame, respectively.
The output is the predicted volume features $\mathcal{V}_{t+n}$ in the future frame at time $n + t$.
The process begins by predicting the absolute flow $\mathcal{F}_t$ for each voxel relative to the current frame coordinates, utilizing an MLPs-based flow decoder.
Subsequently, the voxel flow is transformed into the voxel grid displacement for future frames using the formula $\Delta{t} \cdot n \cdot \mathcal{F}_f$, where $\Delta{t}$ and $n$ are the time interval between adjacent frames, and $n$ denotes the number of frames to be forecasted.
In the nuScenes dataset, $\Delta{t}$ is set to $\frac{1}{12}$ by default, corresponding to the synchronized camera frame rate of 12 frames per second. 
The next step involves transforming the voxel displacement from the current frame to the future frame, relative to the future ego coordinate, by employing the pose transformation matrix.
Finally, the future volume feature is obtained by warping the current volume feature with the predicted voxel displacement.

\begin{table}
\centering
\begin{tabular}{l|c|c|cc} 
\toprule
\multicolumn{1}{c|}{Primitives} & \multicolumn{1}{c}{Variants} & Value & NDS & mAP \\ 
\midrule
\multirow{3}{*}{Mean} & Absolute & - & 26.8 & 25.9 \\ 
\cmidrule{2-5}
 & \multirow{2}{*}{Offset} & \cellcolor{lightgray} 0.25 & \cellcolor{lightgray} 27.3 & \cellcolor{lightgray} 26.5 \\
 &  & ~0.50 & \multicolumn{1}{c}{27.2} & \multicolumn{1}{c}{26.5} \\ 
\cmidrule{1-2}\cmidrule{3-5}
\multirow{3}{*}{Scale} & Fixed & 0.4 & 26.9 & 25.3 \\ 
\cmidrule{2-5}
 & \multicolumn{1}{l|}{\multirow{2}{*}{Learnable}} & \multicolumn{1}{l|}{\cellcolor{lightgray} [0.1, 0.5]} & \multicolumn{1}{l}{\cellcolor{lightgray} 27.3} & \multicolumn{1}{l}{\cellcolor{lightgray} 26.5} \\
 & \multicolumn{1}{l|}{} & [0.2, 0.8] & 26.8 & 26.1 \\ 
\cmidrule{1-1}\cmidrule{2-2}\cmidrule{3-5}
\multirow{2}{*}{Rotation} & Fixed & {[}1, 0, 0, 0] & 27.2 & 26.7 \\
 & \multicolumn{1}{l|}{Learnable} & \cellcolor{lightgray} - & \cellcolor{lightgray} 27.3 & \cellcolor{lightgray} 26.5 \\ 
\bottomrule
\end{tabular}
\caption{Ablation studies on various learning objectives for predicting Gaussian primitives.}
\label{tab:gs_primitive}
\end{table}

\section{Additional Experimental Results}
Due to the space constraint of the main paper, we provide more experimental results here to facilitate a more comprehensive understanding of our proposed method.

\myparagraph{Effects of Gaussian Primitives.}
In \algname, we present autonomous driving scenes as a set of Gaussian primitives $\{g_k = (\mu_k, \Sigma_k, \alpha_k, c_k)\}_{k=1}^K$,  where each Gaussian $g_k$ is parameterized by its 3D position or mean $\mu_k \in \mathbb{R}^3$, a covariance $\Sigma_k$, an opacity $\alpha_k\in [0, 1]$, and spherical harmonics (SH) coefficients $c_k\in \mathbb{R}^k$.
The covariance $\Sigma_k$ is further formulated using a scaling matrix and rotation matrix to ensure its positive semi-definite.
Therefore, it's worth exploring various potential learning objectives for predicting these Gaussian primitives.
To mitigate computational demands during training, this ablation study is confined to the 3D object detection task, employing $50\%$ of the training data during the pre-training phase and $25\%$ during the fine-tuning stage. Results are reported on the entire nuScenes validation set. The model used in this ablation study is consistent with the one employed in the main paper.

The results are presented in Tab.~\ref{tab:gs_primitive}. We designed various experimental variants to explore the effects of different learning objectives. For each variant, we maintained the predictions of other primitives as consistent with those used in the main paper.
We initially exploit the manners for the prediction of Gaussian mean parameters. By default, we transformed the Gaussian mean prediction problem into an offset prediction task using voxel centers as anchors.
We also introduced an offset scaling factor to constrain the range of offset values. Consequently, a Gaussian mean can be obtained by:
\begin{align}
    \mu_k = \mathbf{x}_v+(\mathrm{sigmoid}(\mathcal{O}_k)-0.5)\cdot s,
\end{align}
where $\mathbf{x}_v$ is the coordinate of an anchor point, $\mathcal{O}_k\in \mathbb{R}^{1\times3}$ is the learnable offset, and $s$ is the scaling factor.
We empirically set the $s$ to 0.25, as the same in our main paper, achieving a performance of 27.3 NDS and 26.5 mAP.
When the scaling factor was increased to 0.5, no obvious performance changes were observed.
Furthermore, instead of predicting position offset relative to the centers of volume anchors, we directly regressed the absolute $\mu_k$ via MLPs (second row in Tab.~\ref{tab:gs_primitive}). This resulted in a performance decline in both NDS and mAP, from 27.3 NDS and 26.5 mAP to 26.8 NDS and 25.9 mAP.
This indicates that it's beneficial to predict Gaussian means by predicting the offsets relative to the anchor centers.

In relation to the Gaussian scale and rotation, our primary focus is on exploring the effects of their learnability.
By default, the scale and rotation parameters are set to be learnable, with the scale constrained to the range of $[0.1, 0.5]$. The final scale of Gaussian $k$ is calculated as:
\begin{align}
    S_k = s_l+(s_u-s_l)\cdot \mathrm{sigmoid}(\mathcal{S}_k),
\end{align}
where $s_l$ and $s_u$ represent the lower and upper bounds of the scale, respectively, and $\mathcal{S}_k$ is the learnable scale parameter.
We also experimented with a larger range of $[0.2, 0.8]$, which resulted in a slight decrease in performance.
When the scale and rotation is fixed by setting them as a constant of 0.4 (\ie half of the voxel size) and $[1, 0, 0, 0]$, \ie. the Gaussian degenerates into a spherical form.
Fixing these parameters leads to varying degrees of performance decline, particularly notable when the scale is constant, resulting in a decrease in mAP from 26.5 to 25.3. This indicates that a fixed scale is inadequate for effectively modeling the scene.

\begin{table}[t]
\centering
\small
\begin{tabular}{l|ccc}
\toprule
Methods & Decoder & Memory & Latency \\
\midrule
UniPAD-C & NeRF & 1973MB & 900ms\\
VisionPAD & 3D-GS & 134MB & 70ms\\
\bottomrule
\end{tabular}
\caption{The speed analysis of our method with the UniPAD.}
\label{tab:speed}
\end{table}

\begin{figure*}[t]
\centering
   \includegraphics[width=1.0\linewidth]{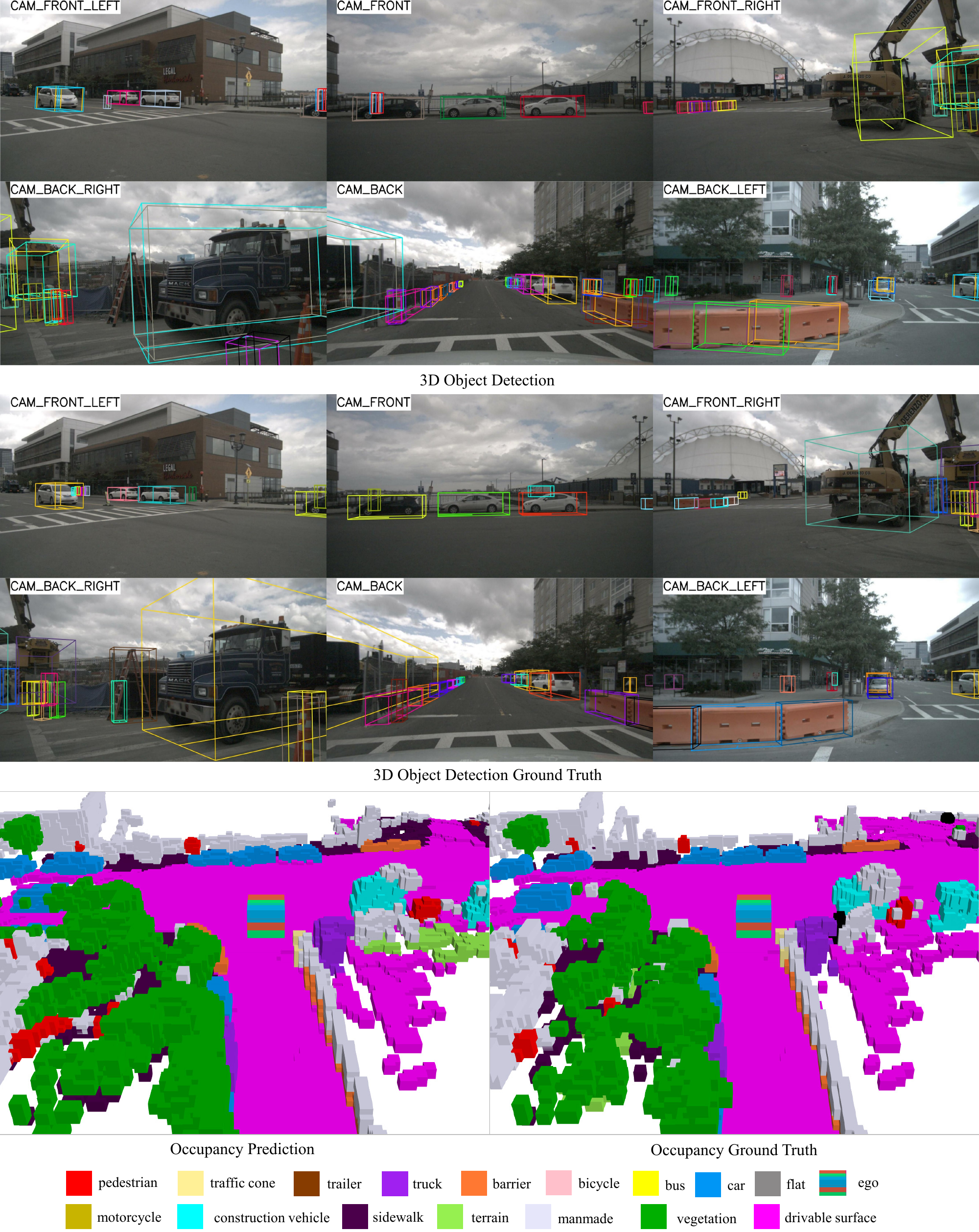}
   \caption{Visualization of 3D object detection and 3D semantic occupancy prediction results. Each detected object instance is depicted by a unique colored 3D bounding box. The legend at the bottom delineates the semantic classes of occupancy.}
\label{fig:vis2}
\end{figure*}

\myparagraph{Speed Analysis.} The NeRF-based methods often suffer from substantial  GPU memory consumption and slow rendering speeds. Furthermore, the resource consumption in NeRF-based methods notably increases with larger rendering sizes.
In contrast, our 3D-GS-based approach maintains faster rendering speeds irrespective of rendering size.
As presented in Tab.~\ref{tab:speed}, we compare the efficiency and memory consumption between UniPAD-C with ours when rendering the same $360\times 640$ images.
We conduct this experiment in a single NVIDIA A100 GPU.
Our approach distinctly results in an approximate 93.2\% reduction in memory usage, alongside an approximate 92.2\% decrease in rendering latency.
This validates the potential of the 3D-GS-based pre-training paradigm.

\myparagraph{UniPAD with V.V. Est. and P.C.} 
To further validate the effectiveness of the proposed photometric consistency (P.C.) and the self-supervised voxel velocity estimation (V.V. Est.), we apply these two strategies in the UniPAD. The results are shown in Tab.~\ref{tab:pc}, our proposed modules achieve significant improvements on UniPAD (+3.3\%/+5.0\%). It is worth noting that 3D-GS renders the \textbf{entire image} with lower computational cost. Compared to \textbf{ray-sampling} approach in UniPAD, 3D-GS increases the effective pixels for photometric consistency supervision, thereby achieving better results.

\begin{table}[h]
\centering
\small
\begin{tabular}{l|c|c|c} 
\toprule
Metrics & UniPAD & w/ V.V. Est. and P.C. & VisionPAD \\ \midrule
NDS & 22.3 & 25.6 & 27.3 \\
mAP & 18.3 & 23.3 & 26.5 \\
\bottomrule

\end{tabular}
\caption{The effectiveness and generalizability of proposed proposed photometric consistency (P.C.) and the self-supervised voxel velocity estimation (V.V. Est.) strategies for pre-training.}
\label{tab:pc}
\end{table}

\myparagraph{Computation Cost of Gaussian Filtering.}
As illustrated in Tab.~\ref{tab:filter}, we demonstrate that using the Gaussian filter can effectively reduce memory usage and improve speed. 
\begin{table}[h]
\centering
\small
\begin{tabular}{l|ccc} 
\toprule
Model & Memory  & Latency & NDS  \\\midrule
w/o filter & 186MB & 90ms & 26.8\\
VisionPAD & 134MB & 70ms & 27.3\\
\bottomrule
\end{tabular}
\caption{The computation cost analysis of Gaussian filtering operation.}
\label{tab:filter}
\end{table}

\section{Additional Visualization Results}
As illustrated in Fig.~\ref{fig:vis2}, we visualize an additional scene to qualitatively assess the efficacy of our method.
To facilitate an intuitive comparison of the perception results, we simultaneously demonstrate the 3D object detection and 3D semantic occupancy prediction within the same scene.
The 3D object detection results are predicted by the UVTR~\cite{li2022uvtr}, while the 3D semantic occupancy prediction results are obtained by the BEVDet~\cite{huang2021bevdet}. Both of them are pre-trained by our method.

From the results, we can not only discern the desirable performance of both 3D object detection and occupancy prediction but also highlight the respective strengths and limitations inherent in these two representations. 
The dense occupancy representation, is capable of depicting arbitrary irregular objects and background obstacles in a detailed voxel format, but tends to demand higher computational resources.
Conversely, 3D bounding boxes primarily represent foreground objects in a sparse format.
In this scene, despite the accurate detections of most objects, some false positive bounding boxes are observed, and the orientation is also not accurate enough to describe the excavator at the front right.
In comparison, the occupancy predictions offer a more satisfactory representation of these critical objects.
Encouragingly, as the visualization results indicate, our pre-training paradigm enhances performance across both tasks.
It further underscores the efficacy and adaptability of our proposed method.

\end{document}